\documentclass{article}
\usepackage{PRIMEarxiv}

\usepackage[utf8]{inputenc} 
\usepackage[T1]{fontenc}    
\usepackage{hyperref}       
\usepackage{url}            
\usepackage{booktabs}       
\usepackage{amsfonts}       
\usepackage{nicefrac}       
\usepackage{microtype}      
\usepackage{lipsum}
\usepackage{fancyhdr}       
\usepackage{graphicx}       
\graphicspath{{media/}}     
\usepackage{amsmath}
\usepackage{microtype}      
\usepackage{cleveref}       
\usepackage{lipsum}         
\usepackage{graphicx}
\usepackage[numbers, sort&compress]{natbib}
\usepackage{doi}
\usepackage{enumerate}
\usepackage{amssymb,enumitem}
\usepackage{multirow}
\usepackage{xcolor}
\usepackage{longtable}
\usepackage{algorithm}
\usepackage{algpseudocode}
\usepackage{array}
\makeatletter
\def\algbackskip{\hskip-\ALG@thistlm}

\usepackage{subcaption}
\title{3D Facial Imperfection Regeneration: Deep learning approach and 3D printing prototypes}

\author{
Phuong D. Nguyen$^{1,}$\thanks{Co-first author.}, Thinh D. Le$^{1,}$\footnotemark[1], Duong Q. Nguyen$^{2}$, Thanh Q. Nguyen$^{1}$, Li-Wei Chou$^{3}$, H. Nguyen-Xuan$^{1, 4,}$\thanks{Corresponding author: CIRTECH Institute, HUTECH University (Email: ngx.hung@hutech.edu.vn).}  \\
$^{1}$CIRTECH Institute, HUTECH University, Ho Chi Minh City, Viet Nam \\
$^{2}$Department of Mathematics and Statistics, Quy Nhon University, Quy Nhon City, Viet Nam\\
$^{3}$Department of Physical Medicine and Rehabilitation, China Medical University Hospital, Taichung, Taiwan \\
$^{4}$Department of Medical Research, China Medical University Hospital, China Medical University, Taichung, Taiwan
}
\begin{document}
\maketitle

\begin{abstract}
This study explores the potential of a  fully convolutional mesh autoencoder  model for regenerating 3D nature faces with the presence of imperfect areas. We utilize deep learning approaches in graph processing and analysis to investigate the capabilities model in recreating a filling part for facial scars. Our approach in dataset creation is able to generate a facial scar rationally in a virtual space that corresponds to the unique circumstances. Especially, we propose a new method which is named 3D Facial Imperfection Regeneration(3D-FaIR) for reproducing a complete face reconstruction based on the remaining features of the patient face. To further enhance the applicable capacity of the present research, we develop an improved outlier technique to separate the wounds of patients and provide appropriate wound cover models. Also, a Cir3D-FaIR dataset of imperfect faces and open codes was released at \url{https://github.com/SIMOGroup/3DFaIR}. Our findings demonstrate the potential of the proposed approach to help patients recover more quickly and safely through convenient techniques. We hope that this research can contribute to the development of new products and innovative solutions for facial scar regeneration.
\end{abstract}


\section{Introduction}
\begin{figure}[!h]
\centering
\includegraphics[scale=0.85]{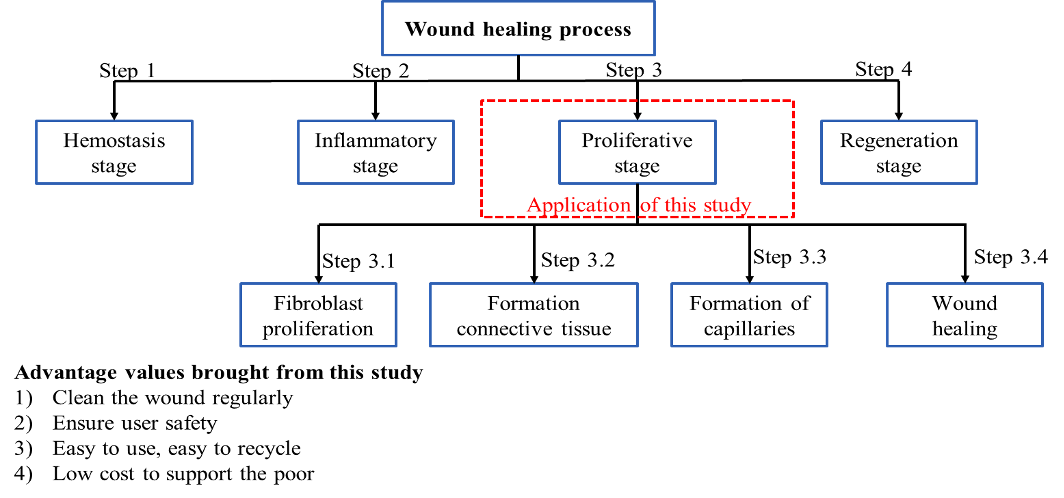}
\caption{The wound healing process of the patient.}
\label{fig:the_wound_healing_process_of_the_patient}
\end{figure}

The field of reconstructing physical flaws in the human body plays significant importance due to its widespread relevance to daily life. Physical disabilities are frequently caused by various factors such as job accidents, traffic accidents, and wars. The presence of disability in the body patient can cause numerous challenges in life, both physically and mentally. One such manifestation is the formation of scars, which are the marks left behind following the healing process of external injuries due to trauma. Except for minor traumas, all wounds originating from accidents, infections, or surgery often produce scars. When the skin is wounded, its recovery and regeneration should meet four stages, including the hemostasis stage, inflammatory stage, proliferative stage, and regeneration stage, as depicted in Fig. \ref{fig:the_wound_healing_process_of_the_patient}. The shape of the scar and how to treat scars are determined by a variety of factors, including the depth, size, and location of the wound as well as the age, sex, and genetics of each patient. Among the steps, step 3 is almost the most important one, which consists of fibroblast proliferation, formation of connective tissue, formation of capillaries, and wound healing. The patient care method must be very attentive throughout this period in order for the wound to heal spontaneously. In accordance with the request of the treating doctor, wounds must be cleaned and disinfected on a regular basis. When the wound dressing process is ineffective, the dressing process may be tough, and it may not be safe to clean the wound. Such an issue becomes problematic and has not been entirely resolved. Scars are unattractive and may lead to physical and emotional symptoms such as itching, pain, sleep disturbance, anxiety, depression, and disruption of daily activities. These psychological sequelae may lead to a lower quality of life. The desire for aesthetic beauty is increasing throughout the world, especially in developing countries. Scars, particularly in areas that cannot be covered by clothing, may affect significantly the life quality of the human.

\begin{figure}[!h]
\centering
\includegraphics[scale=0.20]{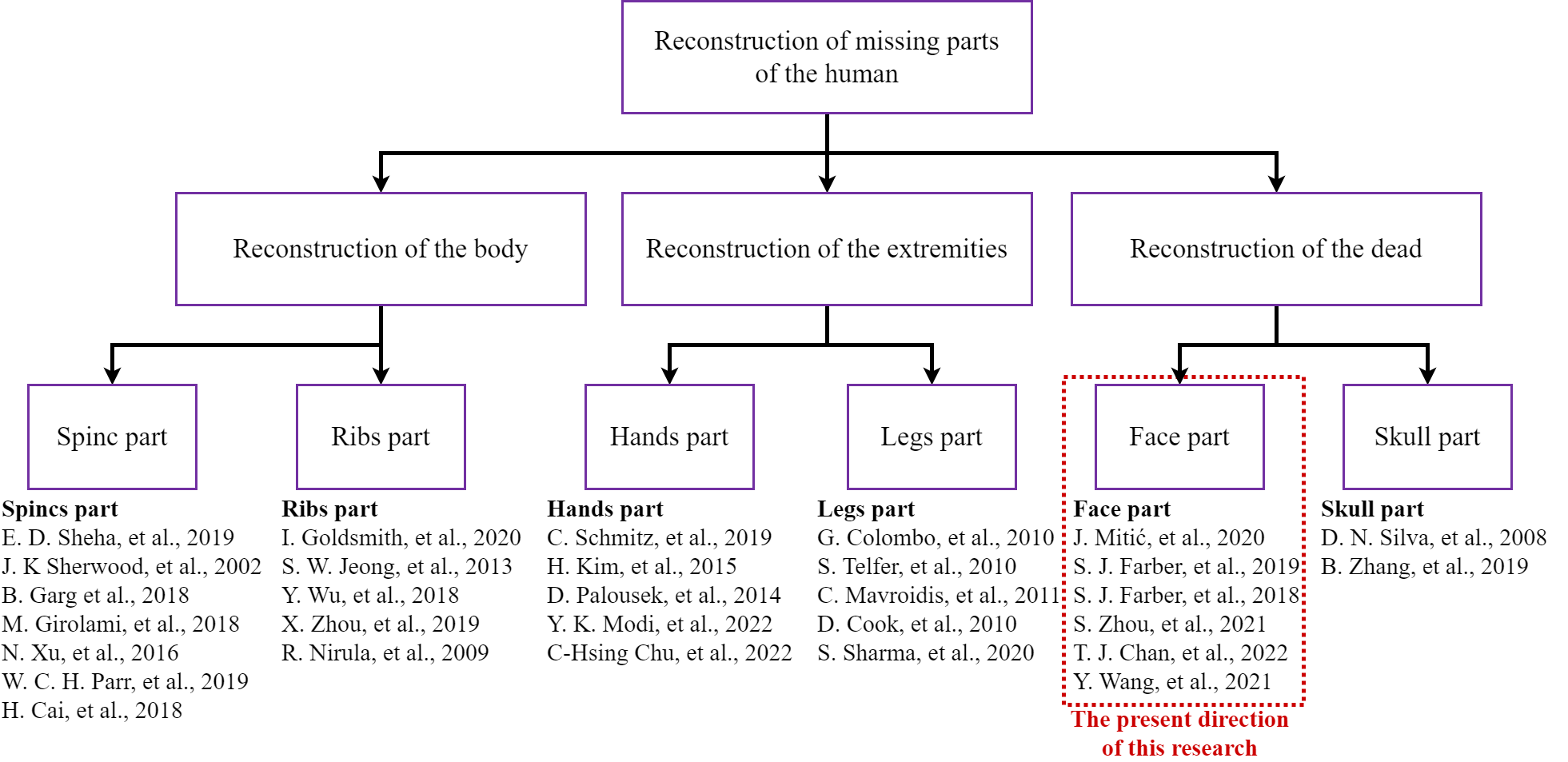}
\caption{Some studies on restructuring the human body.}
\label{fig:some_studies_on_restructuring_the_human_body}
\end{figure}

Of all these commonly injured parts of the human body, the head region at the head and face is the most important and dangerous. Many studies have tried to regenerate damaged structures in this part of the body to help patients recover. As analyzed before, the reconstruction of the human body parts is divided into 3 main research directions \cite{fitch_basic_2008, zhang_biomimetic_2018, yin_tailored_2022}, including body reconstruction \cite{ira_goldsmith_chest_2022, alvarez_design_2021, khullar_prosthetic_2017}, reconstruction of the extremities \cite{poier_development_2021,li_feasibility_2018} and reconstruction of the head as shown in Fig. \ref{fig:some_studies_on_restructuring_the_human_body}. Reconstruction of the extremities has been the subject of numerous publications, particularly with regard to the hands \cite{schmitz_development_2019,kim_case_2015,palousek_pilot_2014,modi_patient-specific_2022,chih-hsing_chu_customized_2022} and legs \cite{colombo_new_2010,telfer_use_2010,mavroidis_patient_2011,cook_additive_2010,sharma_utilization_2020}. Most studies concerned on limb reconstruction due to the relative safety of these areas compared to other parts of the body. In addition, this area is easy to handle and assemble and is less dangerous for the patient. Similarly, reconstruction of the body area is divided into the reconstruction of the spine \cite{sheha_3d_2019,sherwood_three-dimensional_2002,garg_current_2018,girolami_biomimetic_2018,xu_reconstruction_2016,parr_3d_2019,cai_3d_2018} and the rib cage \cite{goldsmith_chest_2020,jeong_conditions_2013,wu_application_2018,zhou_analysis_2019,nirula_rib_2009}. More importantly, two main research directions concerning on the head region consists of facial reconstruction \cite{mitic_reconstruction_2020,farber_reconstructing_2019,farber_face_2018} and skull reconstruction \cite{silva_dimensional_2008,zhang_3d_2019}. Also, injuries to the head can be particularly dangerous and may cause a higher death rate than in other areas. Therefore, research on the reconstruction of the face and skull has still been very challenging. That is a reason why studies on head injuries are also much fewer than injuries to other areas of the body.

Processes to reconstruct virtual structures in the human body are being developed quickly with the help of machine learning models. 3D prototypes for research purposes can be then created by additive manufacturing technologies or 3D printing. The use of machine learning approaches to upgrade 2D models into 3D models of the human facial area is a very potential topic that has been interested in research over the past few decades. Investigating this tough model requires both the recognition algorithm and the correctness of the product after the reconstruction model. The difficulty of the 3D structural reconstruction stems from associated concerns such as the position of facial features, the convexity of the model, and the use of multiple colors to identify the most correct 3D model. In other words, the nature of the problem is heavily influenced by the algorithm model used. Table \ref{tab:1} lists  the key algorithms employed for the reconstruction of the face model.
\begin{table}[!h]
\caption{Some algorithms for the reconstruction of face used in recent years}  \label{tab:1} 
\renewcommand{\arraystretch}{1.5}
\begin{tabular}{|p{1cm}|p{9cm}|>{\centering\arraybackslash}p{3.5cm}|>{\centering\arraybackslash}p{1.2cm}|}
\hline
\centering Period & 
\centering Restrictions& 
\centering External links& 
Sources \\
\hline

\centering Up to 2016 &
\begin{minipage}[t]{\linewidth}
\begin{itemize}[label=-, leftmargin=0.1in,after=\strut]
\item Machine learning models have been widely adopted and proven to be useful in solving various features of detection problems.
\item Several advanced methods have been proposed in the previous states including creating 3D models from signals and recreating 3D models using RGBD scanning. In addition, some applications for recreating human faces, as well as applications that operate on human faces recreated through video.
\item The development during this period is mainly inhibited by the hardware limitations of devices such as smartphones and scanners.
\end{itemize}
\end{minipage}
&
\text{Artificial Neural Network}
\hspace*{-0.27cm} \rule{5.4cm}{0.4pt}
\linebreak
Deep Learning
\hspace*{-0.27cm} \rule{5.4cm}{0.4pt}
\linebreak
Ensemble Methods
\hspace*{-0.27cm} \rule{5.4cm}{0.4pt}
\linebreak
\text{Dimensionality Reduction}
\hspace*{-0.27cm} \rule{5.4cm}{0.4pt}
\linebreak
Bayesian
\hspace*{-0.27cm} \rule{5.4cm}{0.4pt}
\linebreak
Decision Tree
\hspace*{-0.27cm} \rule{5.4cm}{0.4pt}
\linebreak
Regression
&
\cite{agarwal_face_2010, richardson_3d_2016, fangmin_3d_2017, yuan_neural_2002}
\hspace*{-0.28cm} \rule{1.63cm}{0.4pt}
\linebreak
\cite{dou_end--end_2017, sharma_3d_2022, nguyen_fast_2022, dar_real_2022}
\hspace*{-0.28cm} \rule{1.63cm}{0.4pt}
\linebreak
\cite{baek_generative_2020, wang_cascade_2022, liao_video_2019, yaman_comparison_2021, yeping_research_2019}
\hspace*{-0.28cm} \rule{1.63cm}{0.4pt}
\linebreak
\cite{zhang_reconstruction_2004, wang_generalized_2014, wang_auto-encoder_2016, abbad_application_2019}
\hspace*{-0.28cm} \rule{1.63cm}{0.4pt}
\linebreak
\cite{brunton_wavelet_2011, pickup_bayesian_2009, fedorov_robust_2016, claes_bayesian_2010}
\hspace*{-0.28cm} \rule{1.63cm}{0.4pt}
\linebreak
\cite{ashir_facial_2017, pang_face_2005, kazemi_real-time_2014}
\hspace*{-0.28cm} \rule{1.63cm}{0.4pt}
\linebreak
\cite{jackson_large_2017, chinaev_mobileface_2018}
\\
\hline

\centering Up to 2022 &
\begin{minipage}[t]{\linewidth}
\begin{itemize}[label=-, leftmargin=0.1in,after=\strut]
\item New algorithms related to artificial intelligence have been extensively developed, and they are being integrated to create remarkable technological advancements.
\item The hardware systems, including machines that offer high accuracy, convenience, and safety for users, have been developed.
\item Several applications directly interact with the human body with the aid of current 3D printing technology and the development of bio and metal materials.
\end{itemize}
\end{minipage}
&
Deep Learning
&
\cite{zielonka_towards_2022, deng_sub-center_2020, chih-hsing_chu_customized_2022, nguyen_fast_2022, goldsmith_chest_2020, sharma_3d_2022, modi_patient-specific_2022, wang_cascade_2022, yuan_neural_2002, yuchu2020automatic, lequn2021rapid, wei2020deep}
\\
\hline
\end{tabular}     
\end{table}

\begin{figure}[!h]
\centering
\includegraphics[scale=0.22]{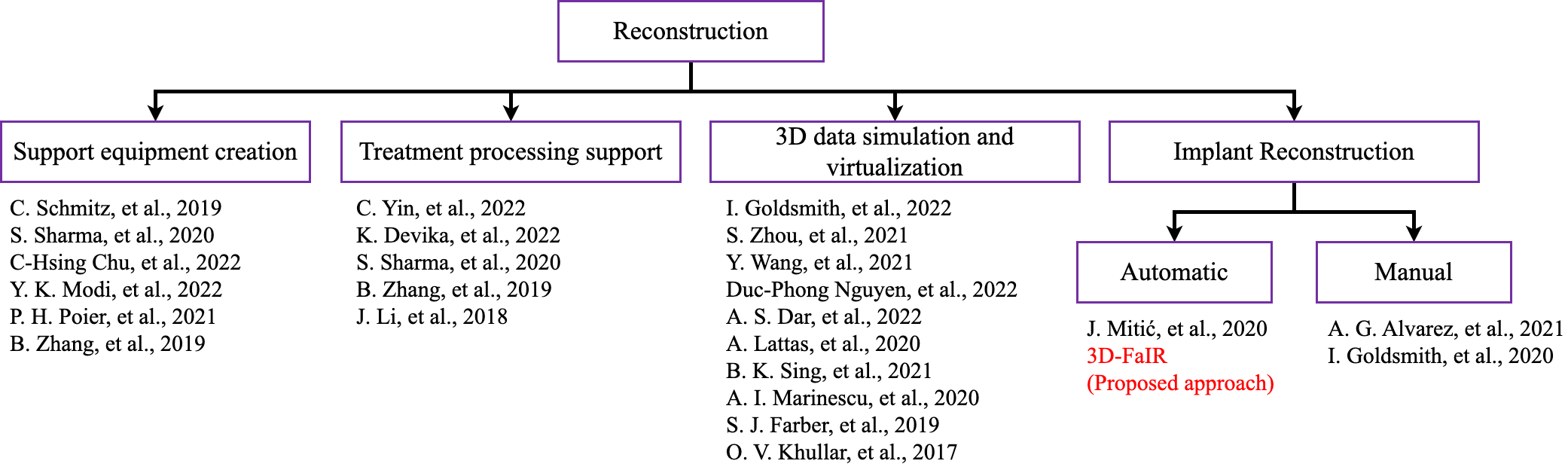}
\caption{Some studies in the direction of 3D reconstruction.}
\label{fig:some_studies_in_the_direction_of_3D_reconstruction}
\end{figure}

As illustrated in Table \ref{tab:1} and Fig. \ref{fig:some_studies_in_the_direction_of_3D_reconstruction}, the relevant studies on the reconstruction of 3D human facial have been extensively devised for a long time. However, the above studies still exist the following gaps:
\begin{itemize}
\item[-] The dataset of faces and facial scars is lacking due to medical research ethics. The ethical considerations in medical research aim to protect the privacy and dignity of human subjects participating in the studies. In the case of facial scars, the inclusion of images of individuals with such scars may lead to stigmatization and discrimination, particularly in employment and social settings. This potential harm to the participants' well-being may outweigh the benefits of having a comprehensive dataset for facial recognition or other related applications. As a result, the lack of a dataset of faces with facial scars may limit the accuracy and effectiveness of facial recognition technologies in identifying individuals with such scars.
\item[-] The facial reconstruction process is still inaccurate due to facial defects because the filling-removal process were designed manually. When addressing facial imperfections, the manual processes of filling and sculpting are typically carried out by a surgical physician. However, Such a manual approach is prone to inaccuracies and inconsistencies, which can lead to suboptimal outcomes for the patient. Additionally, the level of skill and experience of the surgeon can further impact the precision of the procedure, increasing the likelihood of errors. While technology such as computer-aided design (CAD) and 3D printing can partially address these shortcomings, they still require an expert in both the medical and technological fields. Currently, there is no automated method for generating filling material to support the surgeon in the most optimal way.
\item[-] Currently, the majority of research is centered around the manual design of 3D discrete components, with a lack of comprehensive solutions for specific body parts, especially for the facial part. At present, the bulk of research efforts in the realm of 3D design are focused on the manual creation of discrete components as nose, chin, finger joint, ankle bone, etc. Regrettably, there is a paucity of comprehensive solutions that address specific body parts, particularly the all facial region. Despite the increased utilization of 3D printing technologies in the medical field, there remains a significant gap in the development of automated methods for generating filling material that is optimized for facial imperfection correction procedures. This gap highlights a critical need for further research to develop effective techniques for automating the generation of 3D models that accurately capture the intricacies of the facial region. Such advances would be invaluable in streamlining the surgical process, increasing accuracy, and ultimately leading to improved patient outcomes.
\end{itemize}

As a result, to compensate for the deficiencies of earlier studies on this topic, the current work suggests the use of 3D object reconstruction of the  fully convolutional mesh autoencoder  model in the face regeneration problem for trauma patients. The research has exploited advanced machine learning models in image processing and analysis to rebuild facial scars. Facial wound patient models, replicated in virtual space using artificial intelligence, can match true scar states and form more complicated scar models. The findings also show that the proposed method delivers a complete face reconstruction based on the remainder of the face, as opposed to reconstructing by joining one human body component to another. To optimize the present research model, we utilized an improved outlier technique to isolate patients' wounds and generate suitable wound cover models. The obtained results show the potential for developing safe and convenient products to assist patients in their recovery.

\begin{figure}[!h]
	\centering
	\includegraphics[scale=0.16]{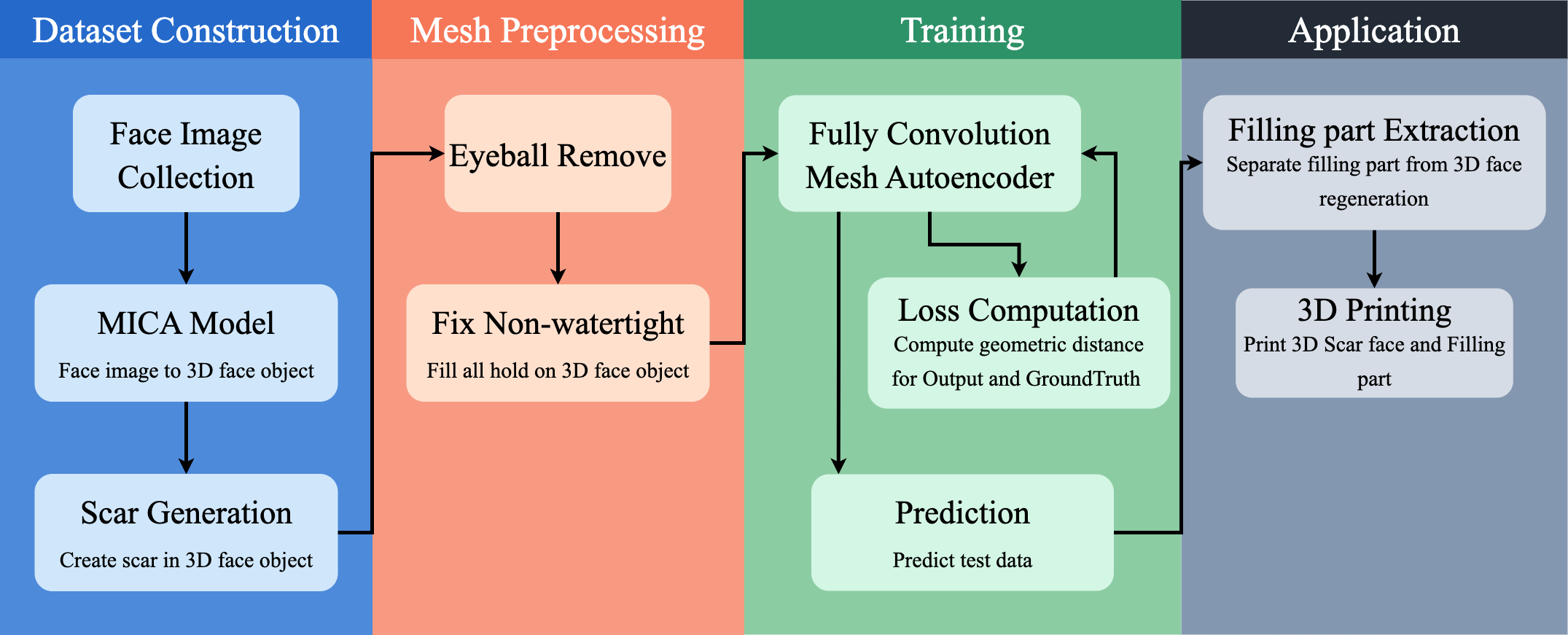}
	\caption{Overview of the data flow process for problem processing.}
	\label{fig:overview_of_the_data_flow_process_for_problem_processing}
\end{figure}

\section{Materials and methods}
\label{sec:headings}
In this section, we present an efficient method to create a human head mesh dataset based on the MICA model by Zielonka et al. \cite{zielonka_towards_2022}. The research direction in the field of defective faces has been severely limited due to the lack of data, mostly caused by patient privacy concerns. To address this issue, we have generated the Cir3D-FaIR dataset using the technique we called Virtual Space. This dataset not only ensures patient privacy but can also be made available publicly to support other researchers. Four main processes including dataset construction, mesh preprocessing, training, and application have been solved. As shown in Fig. \ref{fig:overview_of_the_data_flow_process_for_problem_processing} shows, we have started processing and building the data by collecting the data of the human face image and using the MICA model to get the original 3D face object. Scar Generation is responsible for creating random scars for 3D face objects. Next, in the mesh preprocessing part, we have to refine the data to better fit the Fully Convolution Mesh Autoencoder model such as removing eyeballs and filling holes to fix the non-watertight error. In the Training part, we introduce a new training strategy by changing the calculation object of the loss function. Finally, in the Application part, we offer a method to bring this problem to life by extracting the fill and sending it to 3D printing for further treatment purposes.

\subsection{Generation of dataset}
\label{create_data}
\begin{figure}[!h]
	\centering
	\includegraphics[scale=0.25]{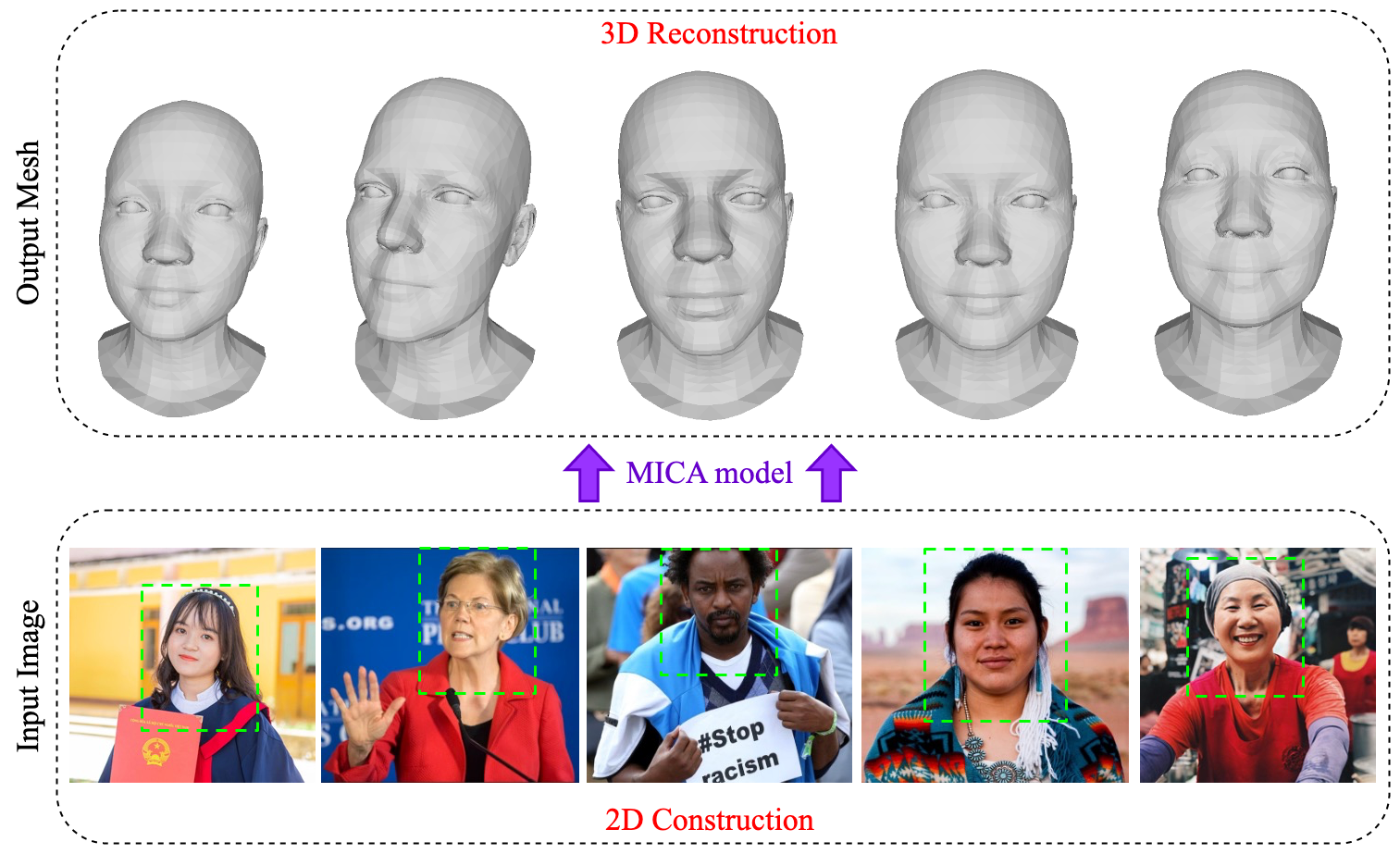}
	\caption{Some input and output of MICA model.}
	\label{fig:some_input_and_output_of_MICA_model}
\end{figure}

For the first phase, we collected photographs of the faces of 3687 individuals from various nations around the world in order  to assure the coverage of facial information of all races. Using the MICA model - Towards Metrical Reconstruction of Human Faces model delivers SOTA - State of the Art results on datasets converted from 2D models to 3D models \cite{zielonka_towards_2022}, these 3687 photographs of human faces were converted to 3D models as displayed in Fig. \ref{fig:some_input_and_output_of_MICA_model}.

\begin{figure}[!h]
	\centering
	\includegraphics[scale=0.25]{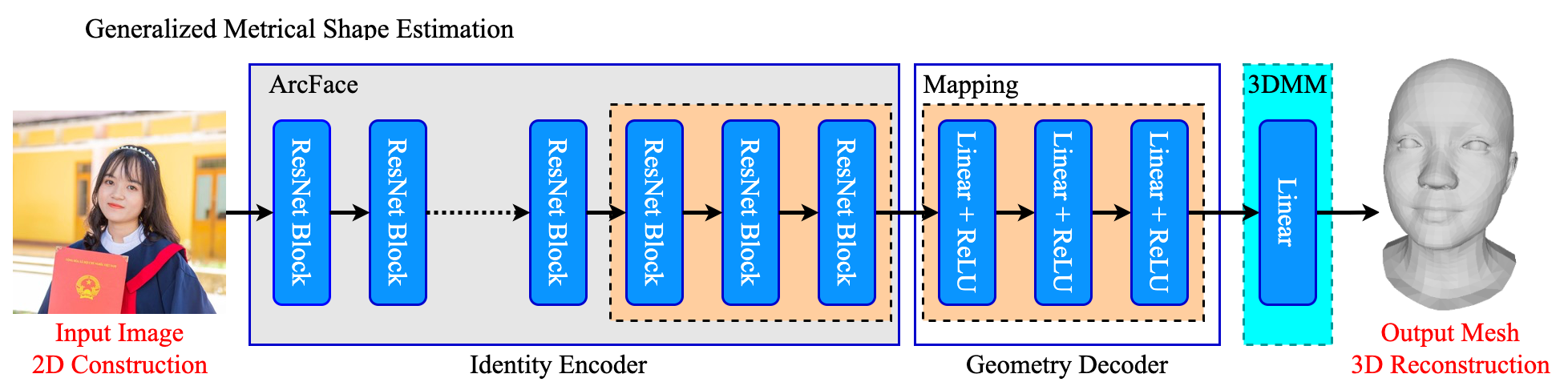}
	\caption{A description of the MICA model.}
	\label{fig:a_description_of_the_MICA_model}
\end{figure}

The MICA model is divided into 2 main parts: Identity Encoder and Geometry Decoder as shown in Fig. \ref{fig:a_description_of_the_MICA_model}. 
\begin{itemize}
	\item[$\bullet$] \textbf{Identity Encoder}: The research adopts the work of Deng et al \cite{deng_sub-center_2020} sub-center ArcFace network in pre-trained facial recognition on large-scale 2D picture datasets. The previous research has revealed the individual characteristics of every single face which can significantly differ in expression, light, and image capability from the Glint360K dataset \cite{an_partial_2021}. According to the present study, the ArcFace sub-center is constructed on the ResNet100 architecture, which is capable of being robust to photographs obtained in surroundings with low stability. Furthermore, the author has extended the architecture of Sub-center ArcFace with a mapping network, which maps the features that Sub-center ArcFace wants to be discernable by their geometry decoder. The network is made up of three fully connected linear hidden layers, and ReLU is used for activation.
	\item[$\bullet$] \textbf{Geometry Decoder}: Since the 3DMM model was too good for facial representation, the author focused on model-based decoder development. In this model, the author uses FLAME \cite{li_learning_2017}, which is a model that uses linear shape space trained from 3800 human head scans, and combines that linear shape space with jaws, necks, eyeballs, pose-dependent corrective blend shapes, and additional global expression blend shapes. In addition, the model was trained at over 33,000 scans and produced more outstanding results than models such as the Face Ware House \cite{cao_facewarehouse_2014} and Basel Face Model \cite{paysan_3d_2009}.
\end{itemize}

\begin{figure}[!h]
\centering
\begin{subfigure}[t]{0.48\textwidth}
  \includegraphics[width=\textwidth,height=6cm]{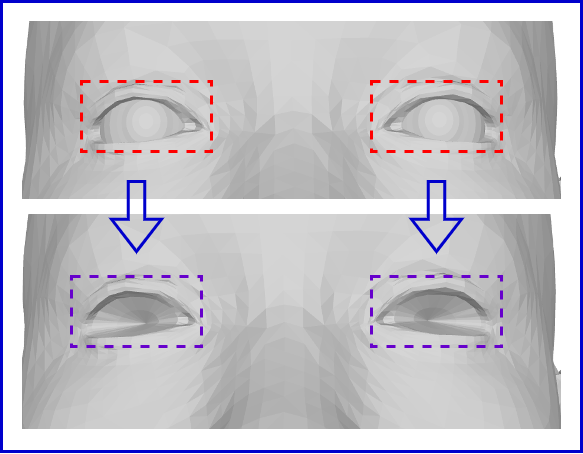}
  \caption{Model filled with holes}
  \label{fig:b_model_of_eyeball_removing}
\end{subfigure}
\begin{subfigure}[t]{0.45\textwidth}
  \includegraphics[width=\textwidth, height=6cm]{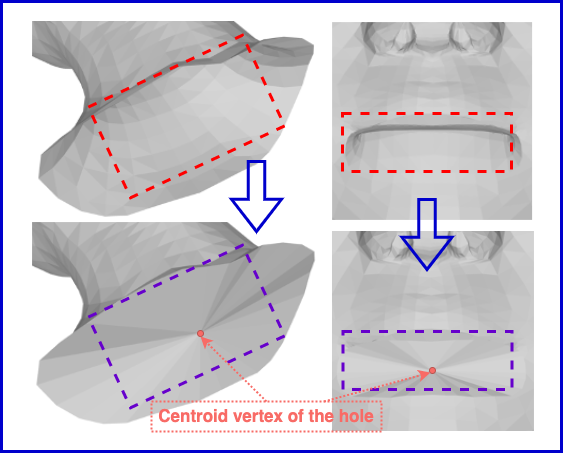}
  \caption{Model of eyeball removing}
  \label{fig:a_model_filled_with_holes}
\end{subfigure}
\caption{Geometrical figures}
\end{figure}

The present study gathers data after using the MICA model, and the replicable model is presented in Fig. \ref{fig:a_description_of_the_MICA_model}. The findings acquired from Fig. \ref{fig:a_description_of_the_MICA_model} are per capita mesh models, and two eyeballs are isolated from the main mesh, yielding a total number of mesh points of 5023. According to the above studies, two eyeballs are not truly required in the human face restructuring model, so this research deleted two eyeballs and the recovery model produced the same findings as Fig. \ref{fig:b_model_of_eyeball_removing}. Because the  fully convolutional mesh autoencoder  model only accepts watertight mesh data \cite{zhou_fully_2020}, this research needs to address the gap that is causing present mesh data to be non-watertight as Fig. \ref{fig:a_model_filled_with_holes}. The proposed method introduces a simpler and easier-to-execute strategy for filling holes in the mesh than earlier studies, including: 1) Find all the edges at the edge hole, so that all the edges in the mesh model are in one face. 2) Connect the edges with corresponding spots, forming sets of holes on the mesh. 3) Take the midway of the pore and connect all the points on the pore to that midpoint, forming a sheet pattern to cover these voids from the extracted points on the edges of the corresponding pores.

\begin{figure}[!ht]
	\centering
	\includegraphics[scale=0.25]{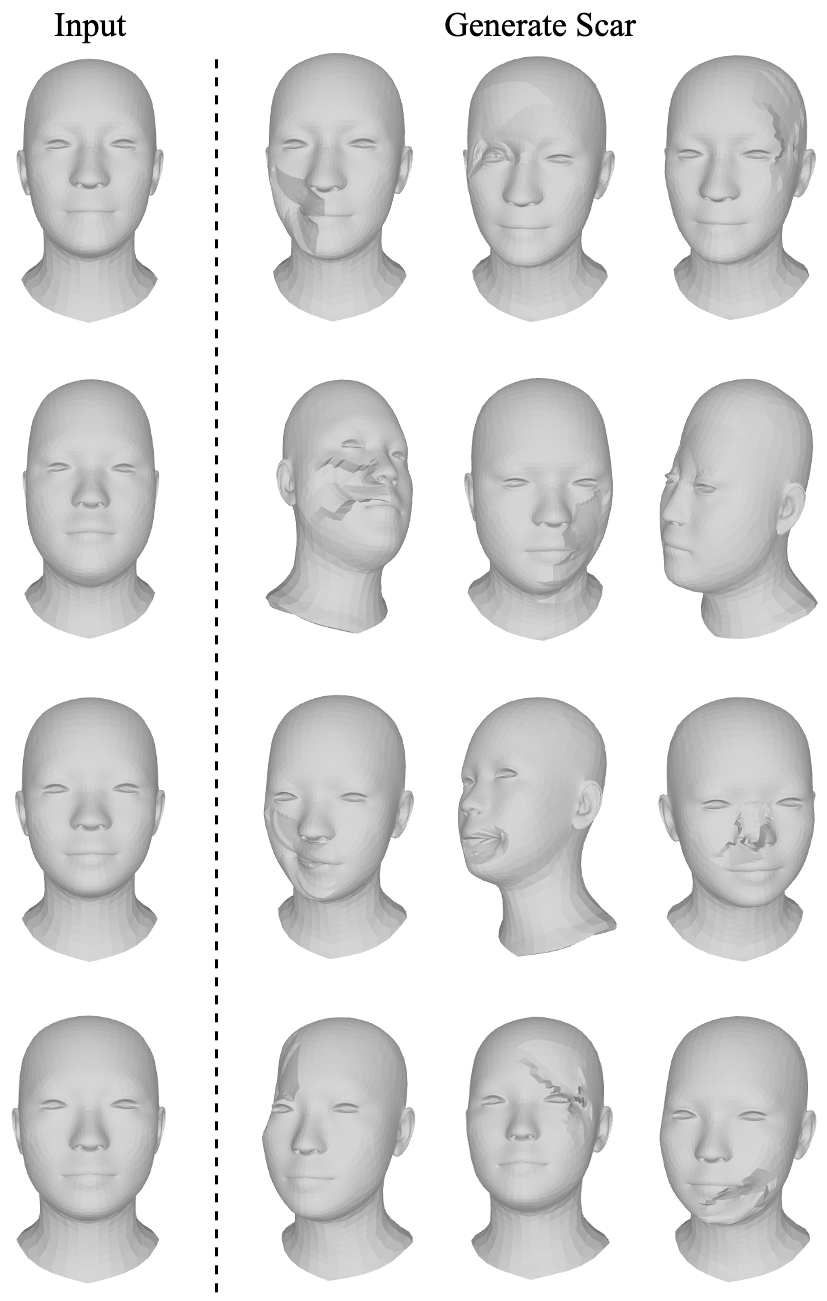}
	\caption{Some simulated models of the actual scarring process.}
	\label{fig:some_simulated_models_of_the_actual_scarring_process}
\end{figure}

The creation of the Cir3D-FaIR dataset involves a multi-step process. First, a simulated MRI scan is created from an actual patient, with the patient's consent, and is then converted into a 3D format. Next, a large number of 3D faces are generated in Virtual Space with MICA dataset, which each product containing different errors in specific parts of the face, such as the eyes, nose, and mouth. This is done to ensure that dataset covers a wide range of possible faces.
Once the 3D faces have been generated, random errors are introduced wherever possible to further increase the variability of dataset. These random errors can include anything from subtle asymmetries to more severe defects. The resulting dataset is intended to be representative of a broad range of facial defects that might be encountered in clinical practice.
Throughout the process of generating the Cir3D-FaIR dataset, medical professionals are consulted to ensure that the dataset is both feasible and useful for research purposes. This includes assessing the accuracy and realism of the simulated defects, as well as evaluating the potential impact of the dataset on patient privacy.
Ultimately, the goal of the Cir3D-FaIR dataset is to provide researchers with a comprehensive and privacy-protected resource for studying facial defects. By generating a diverse and realistic dataset, this research aims to enable further advancements in the diagnosis and treatment of facial abnormalities.
The steps to implement in Virtual Space: 
\begin{enumerate}
	\item Choosing a focal point for facial scars;
	\item Assembling the nearby locations of the scar center specified in step 1 to construct a scar region. The pattern spread out throughout the full scar region generated on the face by adjusting the height of the points in the scar area with the assumption that the scar center is the deepest spot;
	\item At step 3, a concave scar region on the face with the shape, size, and depth of the scar area within a particular range is received after the reconstruction of the scar model from step 2. Random parameters, in particular, are chosen through many trials in order to achieve an empirical visual of the true structure of the actual wound that humans may encounter;
	\item The investigation produces the results shown in Fig. \ref{fig:some_simulated_models_of_the_actual_scarring_process} after employing the strategy presented in this work. The results revealed that while some aspects of the face are deformed, some regions on the face remain fixed and focused on the face. Although the data is not gathered directly from actual patients, we demonstrate that the proposed model is sufficient to assist the research model in simulating and solving real-world problems.
\end{enumerate}

Finally, we modify the structure of the  fully convolutional mesh autoencoder  model by Zhou et al. \cite{zhou_fully_2020} to train on our dataset, allowing the model to reconstruct incomplete faces.

\subsection{Fully Convolutional Mesh Antoencoder}
\label{subsec:FCMA}
The  fully convolutional mesh autoencoder  model by Zhou et al. \cite{zhou_fully_2020} is the basis for training in this study, which aims to address the lack or limitation of using real data for ethical reasons in research. Therefore, this paper presents a new technique, that incorporates convolutional neural networks with graph models, specifically the mesh model.
\subsubsection{vcConv and vcTransConv}
In a regular mesh, if the topology of all neighbors is identical, the learned weight matrix determined for each neighbor vertex can be determined consistently and shared in all vertices of the grid. However, in reality, the vertices are unevenly distributed in space, and each vertex has a different connection and orientation. Therefore, we cannot directly apply the same weighting schemes. Research by Zhou et al. \cite{zhou_fully_2020} has provided a solution to this problem by allowing the weights to vary spatially so that each vertex is free to determine its own convolution weights. At the same time, they introduced two operators, convolution and transposition, denoted \textit{vcConv} and \textit{vcTransConv} respectively, and called the \textquotedblleft coefficients of transformation\textquotedblright~ (\textit{vc}) the local coefficients. These coefficients are in the range of the shared weight basis and are sampled by a set of local coefficients per neighbor. In addition, Zhou et al. \cite{zhou_fully_2020} gave a definition of a Weight Basis (WB), which is a network formed by a virtual space of a defined size on a real-space network, where the real vertices in the mesh within the influence area of a weight basis belongs to that weight basis. The weights in WB are shared across the entire mesh, but the weights of real vertices are calculated inside that weight basis using different functions from vertex to vertex. It was shown that WB is like a pixel in a 2D CNN network, as the model directly computes on WB and then evaluates the weights back for the vertices affected by that weight basis.

Assuming that graph $\mathcal{O}$ is the output sampled from input graph $\mathcal{I}$;  $\mathcal{O}$ has $N$ vertices and each vertex $y_i$ is computed from a local region $\mathcal{N}(i)$ in $\mathcal{I}$; $\mathcal{N}(i)$ has $E_i$ vertices $x_{i, j}, j=1, . ., E_i$. Each convolution layer has one kernel basis defined as $B=\left\{\mathbf{B}_k\right\}_{k=1}^M$, where $\mathbf{B}_k \in \mathbb{R}^{I \times O }$ with local variant coefficients(vc) $A_{i, j}=\left\{\alpha_{i, j, k}\right\}_{k=1}^M, \alpha_{i, j, k} \in \mathbb{R}$. $M$ is chosen close to the average size of a neighborhood. Then, the weight $\mathbf{W}_{i, j}$ on each edge is computed as:
\begin{equation}
\mathbf{W}_{i, j}=\sum_{k=1}^M \alpha_{i, j, k} \mathbf{B}_k,
\end{equation}
and then the convolution is determined by
\begin{equation}
\mathbf{y}_i=\sum_{x_{i, j} \in \mathcal{N}(i)} \mathbf{W}_{i, j}^T \mathbf{x}_{i, j}+\mathbf{b}.
\end{equation}
where $\mathbf{x}_{i, j}\in \mathbb{R}^{I}$ are the input feature; $\mathbf{b}$ is the learned bias; $A_{i, j}$ are different for each vertex $x_{i, j}$ in $\mathcal{N}(i)$ of each $y_i$, but $B$ is shared globally in one layer. The parameters used in this approach are learnable and shared across the entire dataset. 
\subsubsection{vdPool, vdUnpool, vdUpRes, and vdDownRes}
To preserve the functionality of traditional Convolutional Neural Networks (CNNs), Pool and Unpool Layers are commonly used. Common types of pooling include max pooling and average pooling. Given a mesh with $n$ vertices $V = {v_1, v_2, \ldots, v_n}$ and their corresponding features $F = {f_1, f_2, \ldots, f_n}$. Two pooling layers, maximum and average can be defined, respectively, as follows
\begin{equation}
f_{i,~maxpool} = \max_{j \in \mathcal{N}(i)} f_j,
\end{equation}
\begin{equation}
f_{i,~avgpool} = \frac{1}{|\mathcal{N}(i)|}\sum_{j \in \mathcal{N}(i)} f_j,
\end{equation}
where $\mathcal{N}(i)$ represents the set of neighboring vertices of vertex $i$ in the mesh and $|\mathcal{N}(i)|$ is the number of neighboring vertices. This creates a coarser representation of the original mesh with fewer vertices, as each pooled vertex represents the average or maximum feature value of the neighborhood. The corresponding unpooling layer restores the original resolution of the mesh by copying the pooled features to their corresponding vertices. However, max or average pooling can be less effective when applied to grids with uneven peak density because they treat all regions of the input grid equally, regardless of their importance. This can result in important features being averaged out or overlooked, leading to a loss of information and a decrease in performance. To address this issue, various pooling methods have been proposed that take into account the local density of peaks or the spatial relationships between features. Thus, Zhou et al. \cite{zhou_fully_2020} described a technique for feature aggregation using Monte Carlo sampling. The approach is inspired by the work of Hermosilla et al. \cite{hermosilla2018monte}, who estimated the vertex density using 3D coordinates of neighboring vertices. However, in more general cases, such information may not be available for each layer. In \cite{zhou_fully_2020}, instead of designing a density estimation function, the authors propose to learn the optimal variant density coefficients (\textit{vd}) across all training samples. These coefficients are defined per node after pooling or unpooling, and they allow the network to adaptively adjust the density of features for each node. By learning the optimal vd coefficients, the network can better capture the important features of the input and improve performance. It is worth noting that vd is defined per node after pooling or unpooling. In \cite{zhou_fully_2020}, vdPool and vdUnpool layers act as pooling and unpooling layers in CNN,, respectively. Accordingly, each local vertex $j$ has a density coefficient is defined as follows
\begin{equation}
\rho_{i, j}^{\prime}=\frac{\left|\rho_{i, j}\right|}{\sum_{j=1}^{E_i}\left|\rho_{i, j}\right|},
\label{pool}
\end{equation}
where $\rho_{i, j} \in \mathbb{R}$ are training parameters, shared across the dataset. Specifically, Eq. (\ref{pool}) is an aggregate function used in vdPool and vdUnpool layers. Due to the $vd$ coefficient normalization, the vdPool and vdUnpool do not perform any rescaling nor change the mean values of the input feature map.
Similarly, the residual layer is written as:
\begin{equation}
\mathbf{y}_i=\sum_{x_{i, j} \in \mathcal{N}(i)} \rho_{i, j}^{\prime} \mathbf{C x}_{i, j}.
\end{equation}
If the input and output feature dimensions are identical, then $\mathbf{C}$ represents an identity matrix. However, if the input and output feature dimensions differ, then $\mathbf{C}$ is a learned $O \times I$ matrix that is learned and shared across all graph nodes.
\begin{figure}[!h]
	\centering
	\includegraphics[scale=0.038]{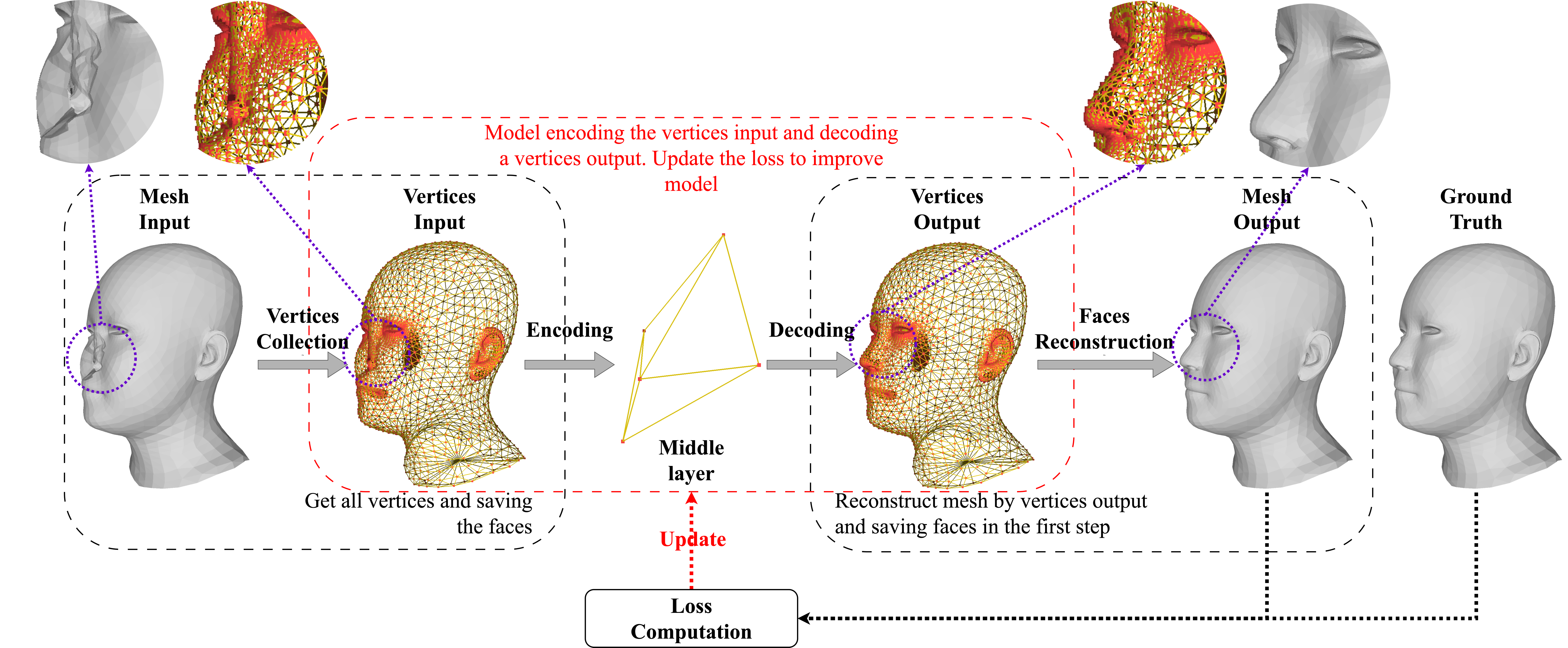}
	\caption{The coaching model we recommend.}
	\label{fig:the_training_model_we_recommend}
\end{figure}
According to this layer, a residual block is used for up-and-down sampling. At this point, the input is run through vcConv for down-sampling or vcTransConv for up-sampling and then the data is run through the activation layer function.

We conduct the analysis to choose the appropriate activation function for the problem. The Rectified Linear Unit (ReLU) and the Exponential Linear Unit (ELU) are both popular activation functions used in neural networks. They are both non-linear functions that introduce non-linearity into the output of a neural network, which is necessary for the network to learn complex patterns in the data. The ReLU function is defined as:
$$f(x) = \max(0, x),$$
where $x$ is the input to the activation function. The ReLU function is a piecewise linear function that returns zero for negative inputs and the input value itself for positive inputs. The ReLU function has several advantages, including its simplicity, computational efficiency, and sparsity. The ELU function, on the other hand, is defined as:
\begin{align}
f(x) = \begin{cases}
x & \text{if } x > 0,\\
\alpha (\exp(x) - 1) & \text{if } x \leq 0,\\
\end{cases}
\end{align}
where $\alpha$ is a hyperparameter that controls the value of the function for negative inputs. The ELU function is a smooth and non-monotonic function that returns the input value for positive inputs and has an exponential component for negative inputs. The ELU function has several advantages, including its smoothness, differentiability, and small negative output for negative inputs. One key advantage of the ELU function over the ReLU function is its ability to avoid the \textquotedblleft dead ReLU\textquotedblright~problem. In some cases, the ReLU function can cause neurons to become permanently inactive if their output falls below zero. This can lead to a reduction in the expressive power of the network and lower performance. The ELU function avoids this problem by having a non-zero output for negative inputs, which can help to prevent neurons from becoming inactive. Another advantage of the ELU function is its small negative output for negative inputs. This can help to improve the robustness of the network to noise and outliers, which is important in many applications. 

From the above analysis, we can see that ELU is roughly smooth while ReLU is completely smooth. Therefore, ReLU is chosen as the activation function for this problem. The output from the ELU activation layer is combined with the output from the vdDownRes or vdUpRes layer and produces the final output. The implementation process is shown in Figure  \ref{fig:the_training_model_we_recommend}.

\subsection{Training strategy}
\label{sec:model}
The selection of an appropriate loss function is essential in many machine learning problems, including mesh reconstruction. Let $v_{i}(x,y,z) \in \mathcal{M}_{in}(V,E,F)$ and $v'_{i}(x',y',z') \in \mathcal{M}_{out}(V', E',F')$ be the vertices in the input and output mesh, respectively. In  \cite{zhou_fully_2020}, the loss function is calculated as the average of the sum of Euclidean distances between the corresponding vertices in the input mesh and output mesh. Accordingly, the loss function is written as follows:
\begin{equation}
d(v_i, v'_i)= ||v_i - v'_i||_2.
\label{euclid}
\end{equation}
Therefore, the loss function can be expressed as
\begin{equation}
\mathcal{L}(\mathcal{M}_{in}, \mathcal{M}_{out}) = \dfrac{1}{n} \sum_{i=1}^{n} d(v_i, v'_i),
\end{equation}
where $n = n(E) = n(E')$ is the number of vertices.

Zhou et al \cite{zhou_fully_2020} described a training model in which both the input and the comparison value are used to compute the output from the model. This approach helps the model to learn similar to the structure of the input mesh. However, our goal is to train a model to locate the wound and reconstruct it to resemble a natural face. Therefore, this loss function needs to be justified. To solve this problem, we introduce the loss function by modifying the training architecture of the fully convolutional mesh autoencoder model.

According to Subsection \ref{create_data}, we get 3687 ground truth mesh faces of people before the injury. In the problem of this research, the input and the ground truth are two different meshes. Thus, the inclusion of the input is a human head mesh with wounds and the ground truth is a human head mesh before the injury, and then the output generated from the model is calculated with that ground truth. Similar to $\mathcal{M}_{in}$,  $\mathcal{M}_{out}$, let $v''_i(x'',y'',z'') \in \mathcal{M}_{gt}$ be the vertices in the ground truth mesh. Then, the loss function is adjusted as follows:
\begin{equation}
\mathcal{L}(\mathcal{M}_{out}, \mathcal{M}_{gt}) = \dfrac{1}{n} \sum_{i=1}^{n} d(v'_i, v''_i) = ||v'_i - v''_i||_2,
\end{equation}
where $d(v'_i, v''_i)$ is the Euclidean distance of the vertices on $\mathcal{M}_{out}$ and $\mathcal{M}_{gt}$; $n = n(E') = n(E'')$ is the number of vertices. With this correction, the model can learn how to repair the damage in a way that brings the injured face back to its original state, rather than simply filling the wound. The proposed training model from this research is exhibited in Fig. \ref{fig:the_training_model_we_recommend}.

\begin{figure}[!h]
	\centering
	\includegraphics[scale=0.35]{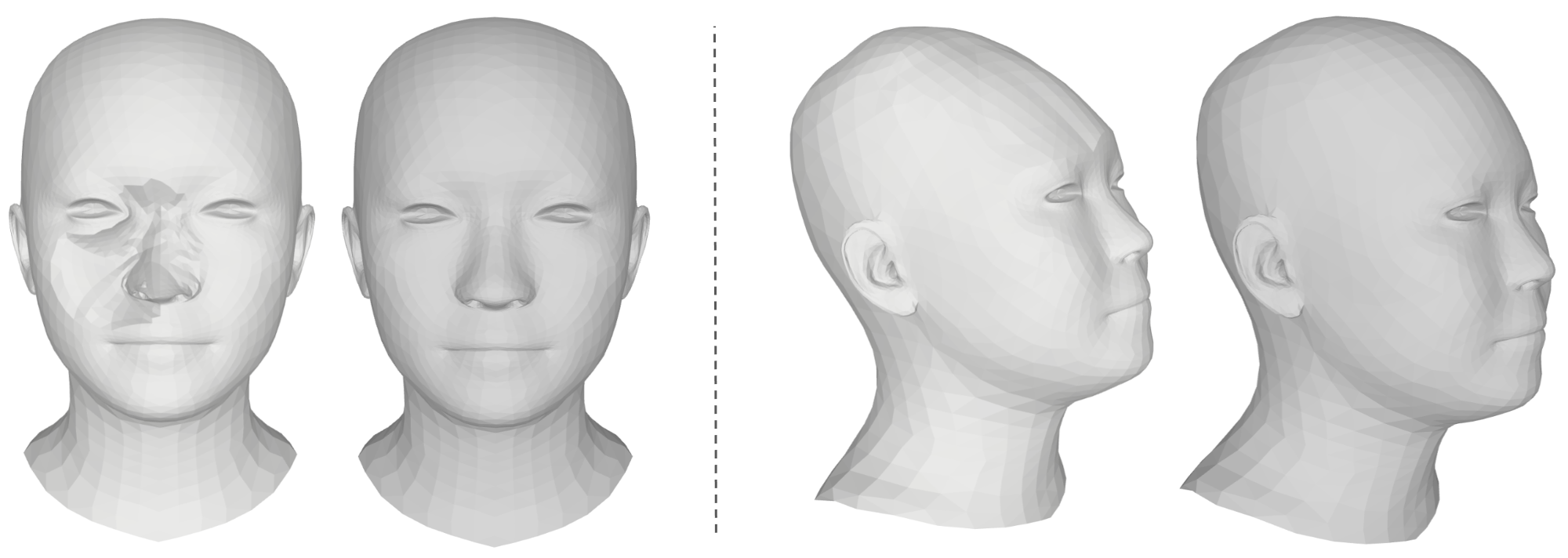}
	\caption{Results obtained from the proposed model.}
	\label{fig:results_obtained_from_the_proposed_model}
\end{figure}

\subsection{Filling extraction}
\begin{figure}[!h]
	\centering
	\includegraphics[scale=0.35]{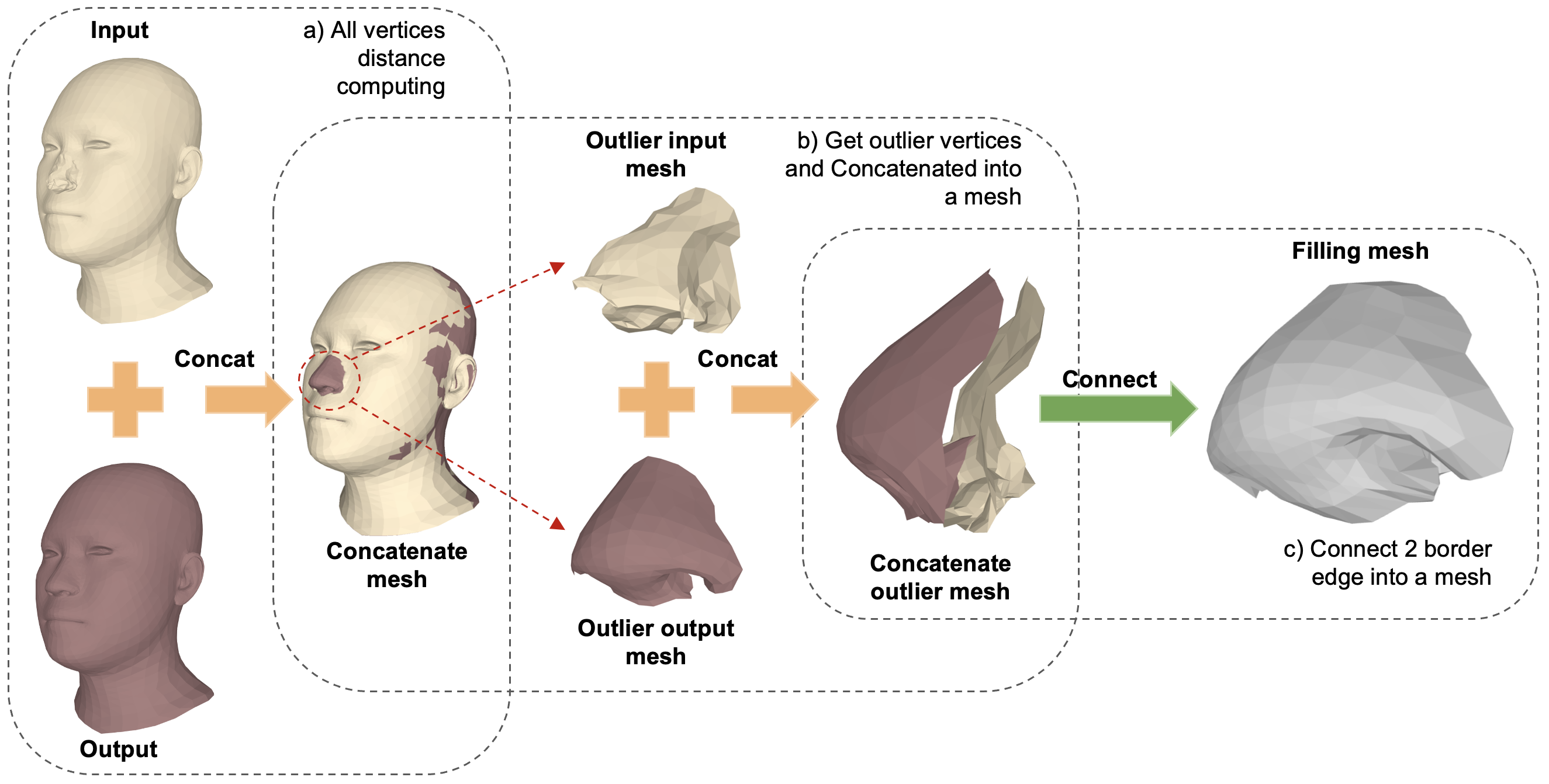}
	\caption{The procedure for the wound filling part.}
	\label{fig:the_procedure_for_the_wound_filling_part}
\end{figure}

Fig. \ref{fig:results_obtained_from_the_proposed_model} showed results derived from the training model in Section \ref{sec:model}. For the surgery of patients, the wound filling part can be then reproduced by the use of 3D printing technology. We introduce a new technique called \textquotedblleft filling extraction\textquotedblright~ which has been implemented to fill in the missing parts with high accuracy and low computational cost. In particular, the input mesh and the ground truth mesh are both created from the same original mesh, as given in Section \ref{create_data}. The output mesh tends to reform like the ground truth mesh. Therefore, the model can take the output mesh \textquotedblleft minus\textquotedblright ~the input mesh to obtain the filling part of the wound on the face. Because the output mesh is generated from a deep learning model, it itself exists certain errors, which need to be fixed to achieve the most accurate model. Therefore, when taking the output mesh \textquotedblleft minus\textquotedblright~ the input mesh, in addition to filling the wound area, the wrong parts are also removed according to the concatenate mesh as illustrated in Fig. \ref{fig:the_procedure_for_the_wound_filling_part}a. To remedy this, the outlier method was used to eliminate the underlying errors. It is noted that the  fully convolutional mesh autoencoder  \cite{zhou_fully_2020} model is characterized by the fact that the output and the input have the same number of points and different point order. Accordingly, we concatenate both the input and the output in three-dimensional space. We assume that $D$ is the set of distances of $v_{i}(x,y,z) \in \mathcal{M}_{in}(V,E,F) \text{ and } v'_{i}(x',y',z') \in \mathcal{M}_{out}(V', E',F')$ described by:
\begin{equation}
D := \left\lbrace d(v_i, v'_i)\right\rbrace ^{n}_{i=1} = \left\lbrace ||v_i - v'_i||_2 \right\rbrace ^{n}_{i=1} = \left\lbrace \sqrt{(x_i - x'_i)^2 + (y_i - y'_i)^2 + (z_i - z'_i)^2} \right\rbrace ^{n}_{i=1}
\label{eq_D}
\end{equation}
where $n$ is the number of vertices of input and output meshes. We calculate the mean $\mu_{D}$ and variance $\sigma_{D}$ from the set $D$, respectively, as follows
\begin{equation}
\mu_D = \frac{1}{n} \sum_{i=1}^{n} D_i, \quad \sigma_{D} =  \sqrt{\frac{1}{n}\sum_{i=1}^{n} (D_i - \mu_{D})^2},
\label{eq_mu}
\end{equation}
where $n$ is the number of elements in the set $D$, and $D_i$ is the $i$-element of the set $D$. We define the outlier extraction formula to show the error correction as follows
\begin{equation}
D_{outlier} := \left\lbrace (v_i, v'_i) \mid |d(v_i, v'_i)-\mu_{D}| > 2\sigma_{D}\right\rbrace,
\label{eq_outlier}
\end{equation}
where $1<i<n$. From $D_{outlier}$, we get the coordinates of the outliers on the input mesh and output mesh respectively. The outliers are then concatenated into space, which gets the same mesh as the concatenate outlier mesh in Fig. \ref{fig:the_procedure_for_the_wound_filling_part}. Since the concatenate outlier mesh is still two separated mesh parts, we connect them by staying at the outer edges. Then, we take additional vertices from the output mesh that are directly connected to each vertex at the outer edge of the outlier output mesh. Then, the calculation joins the outlier input mesh and the newly added vertices of the outlier output mesh together with the result eventually formed a filling mesh for the fig-like wound as shown in Fig.  \ref{fig:the_procedure_for_the_wound_filling_part}.

\section{Results and discussion}
\begin{figure}[!h]
	\centering
	\includegraphics[scale=0.32]{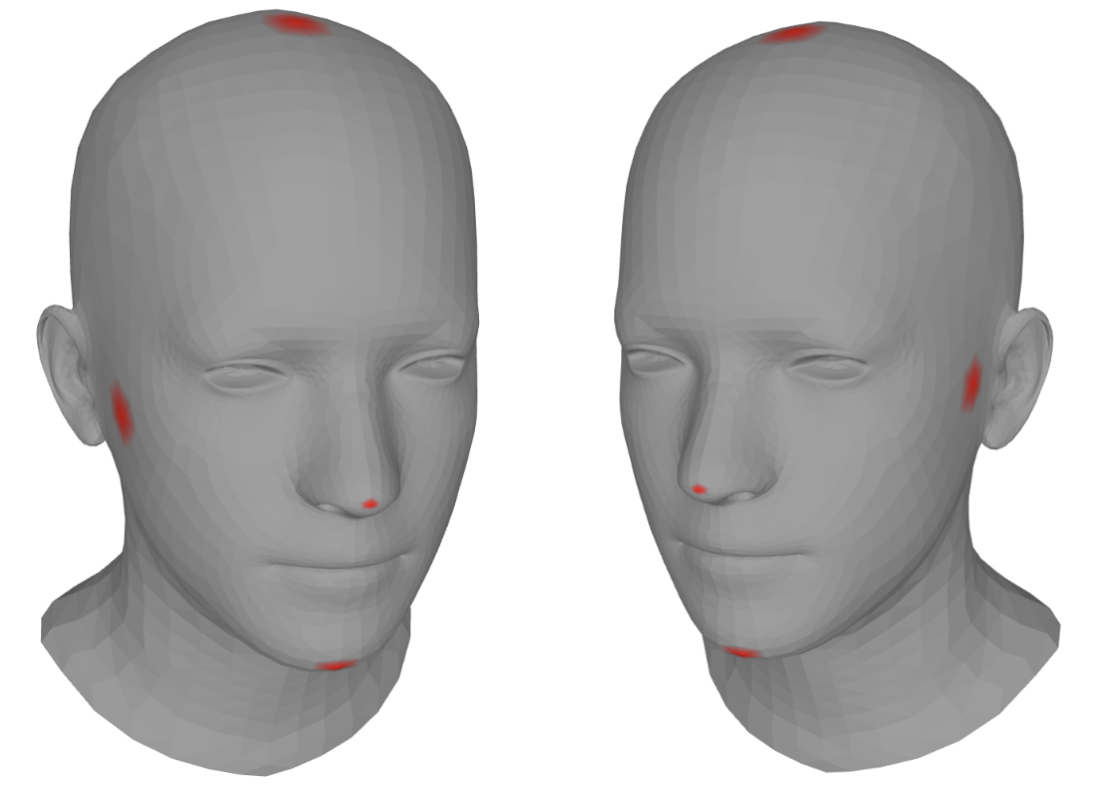}
	\caption{Illustration for the middle layer of the model.}
	\label{fig:illustration_for_the_middle_layer_of_the_model}
\end{figure}

\begin{figure}[!h]
	\centering
	\includegraphics[scale=0.32]{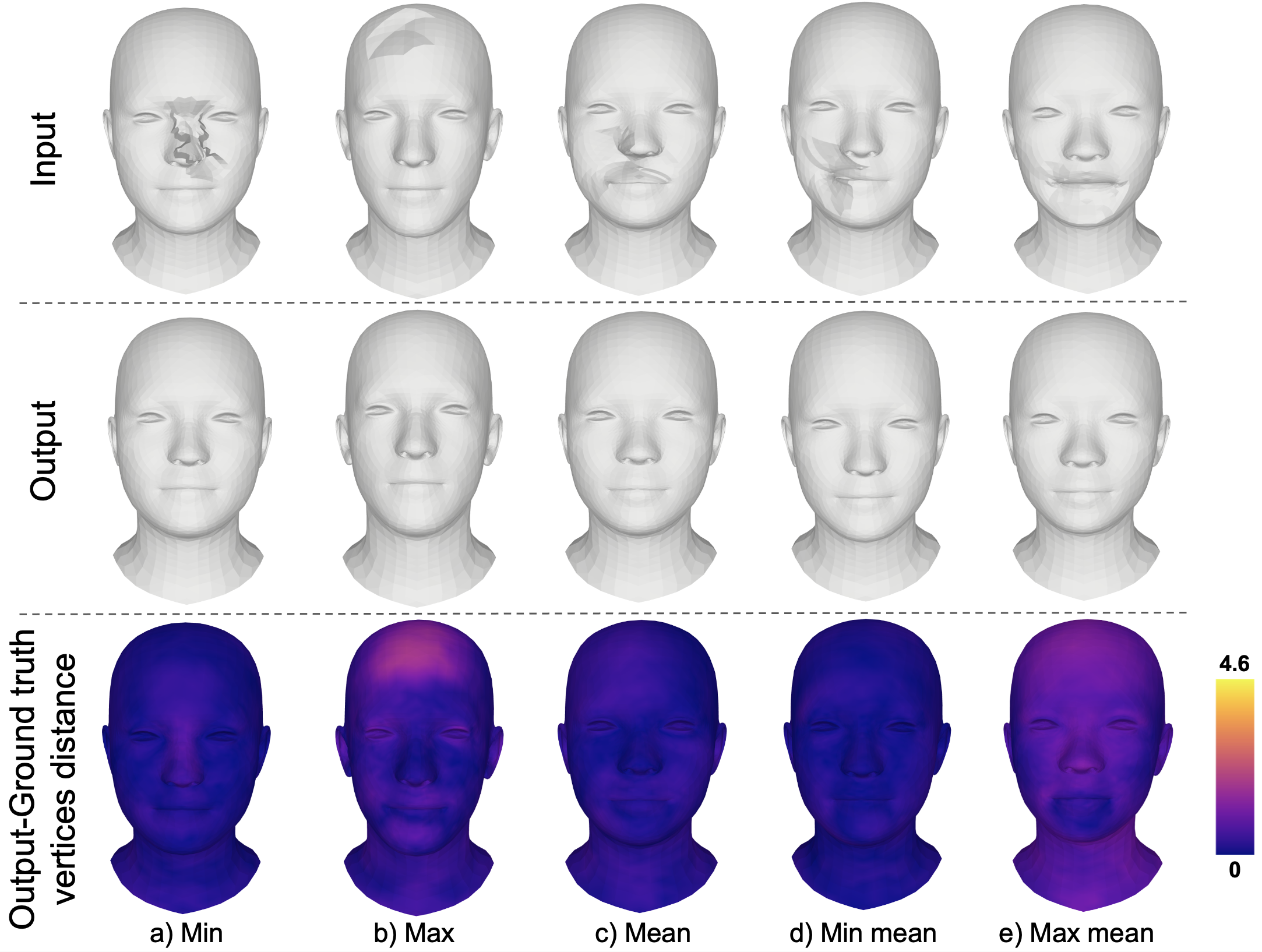}
	\caption{The results are statistically extracted from the model with the test suite (3678 meshes).}	                           \label{fig:the_results_are_statistically_extracted_from_the_model}
\end{figure}

\begin{figure}[!h]
	\centering
	\includegraphics[scale=0.48]{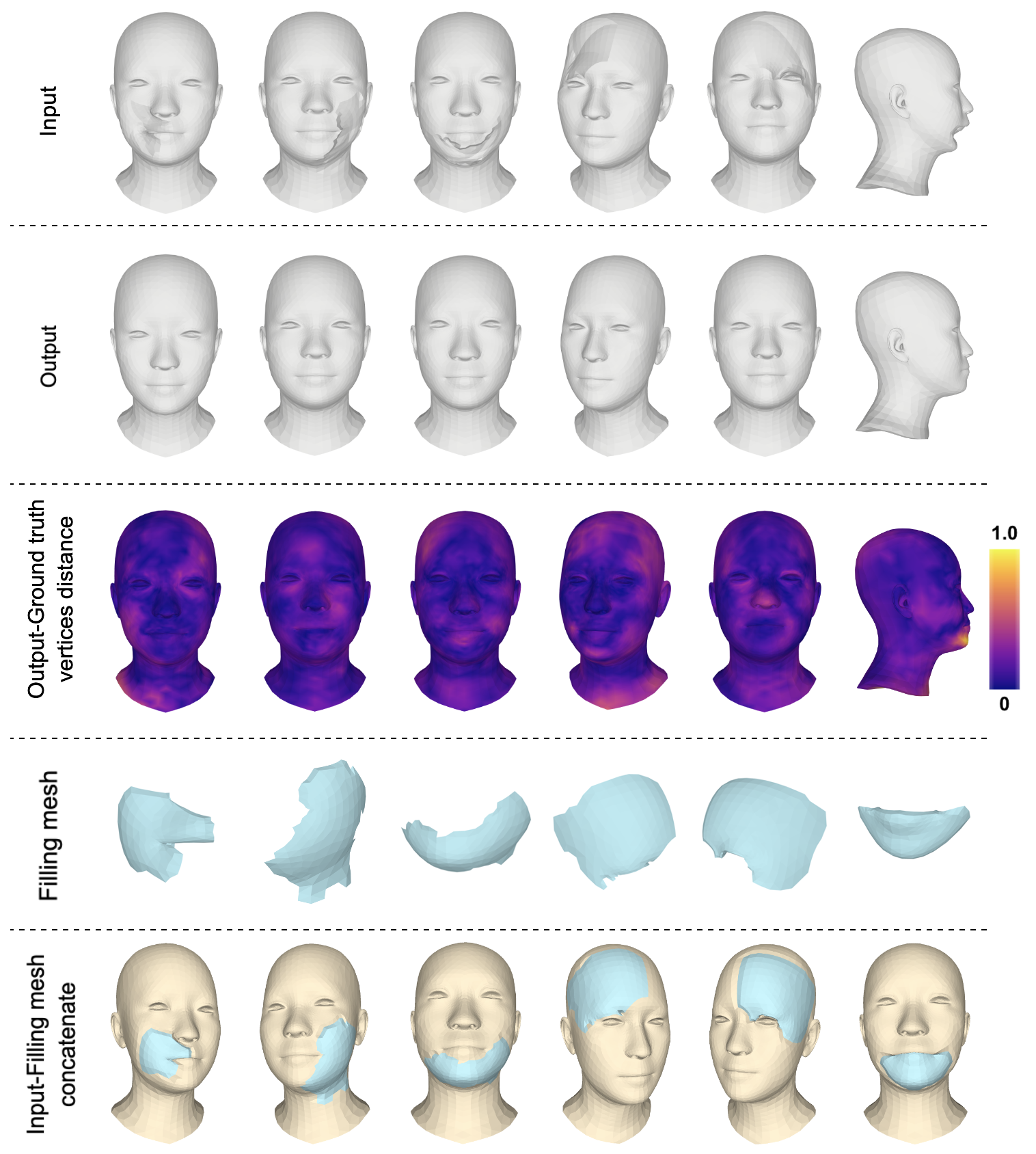}
	\caption{The results were randomly extracted from the model with the test suite (3678 meshes).}
	\label{fig:the_results_were_randomly_extracted_from_the_model_with_the_test_suite}
\end{figure}

\begin{figure}[!h]
	\centering
	\includegraphics[scale=0.15]{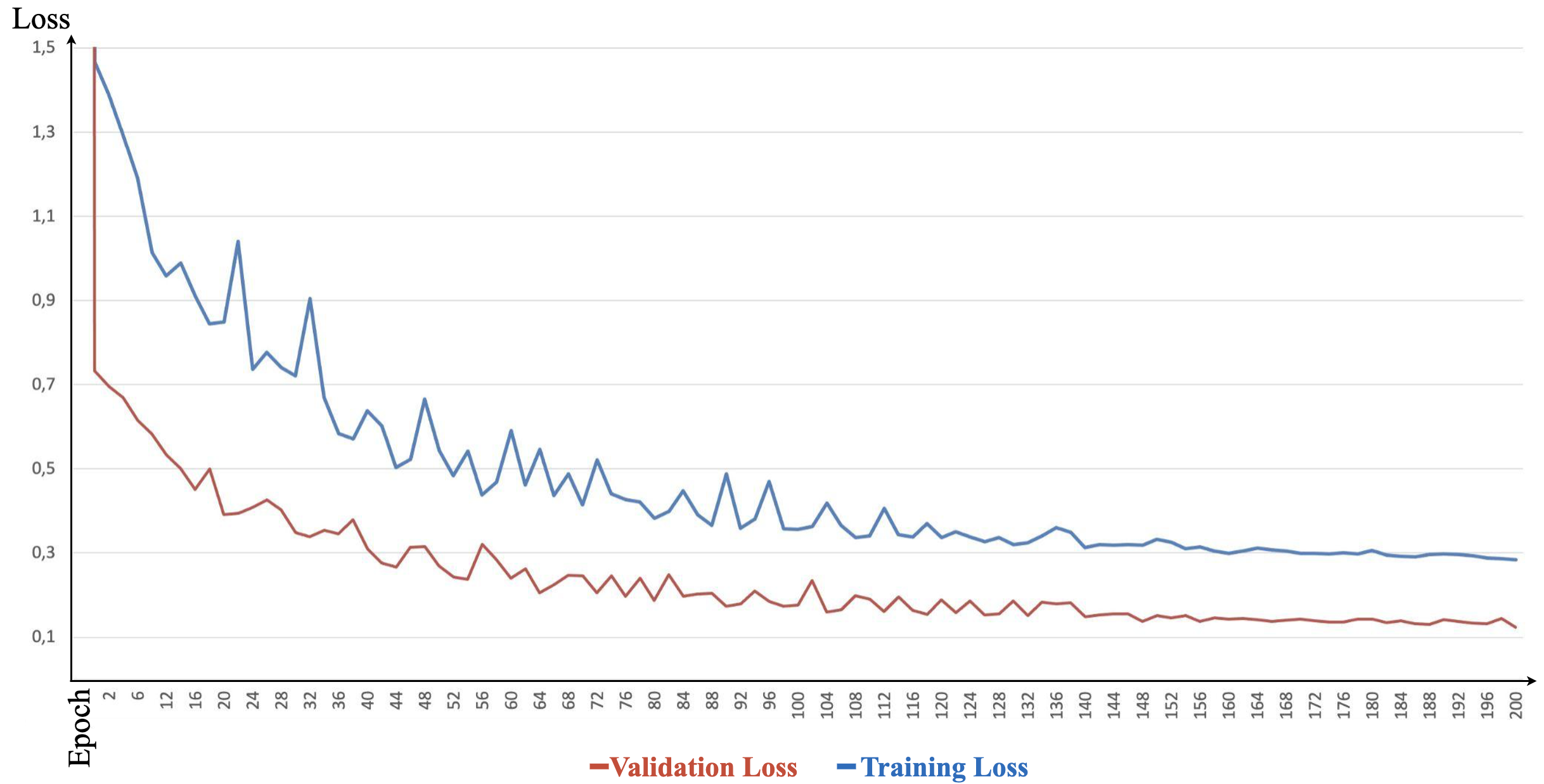}
	\caption{The loss function value corresponding to each Epoch.}
	\label{fig:the_loss_function_value_corresponding_to_each_epoch}
\end{figure}

\begin{figure}[!h]
	\centering
	\includegraphics[scale=0.16]{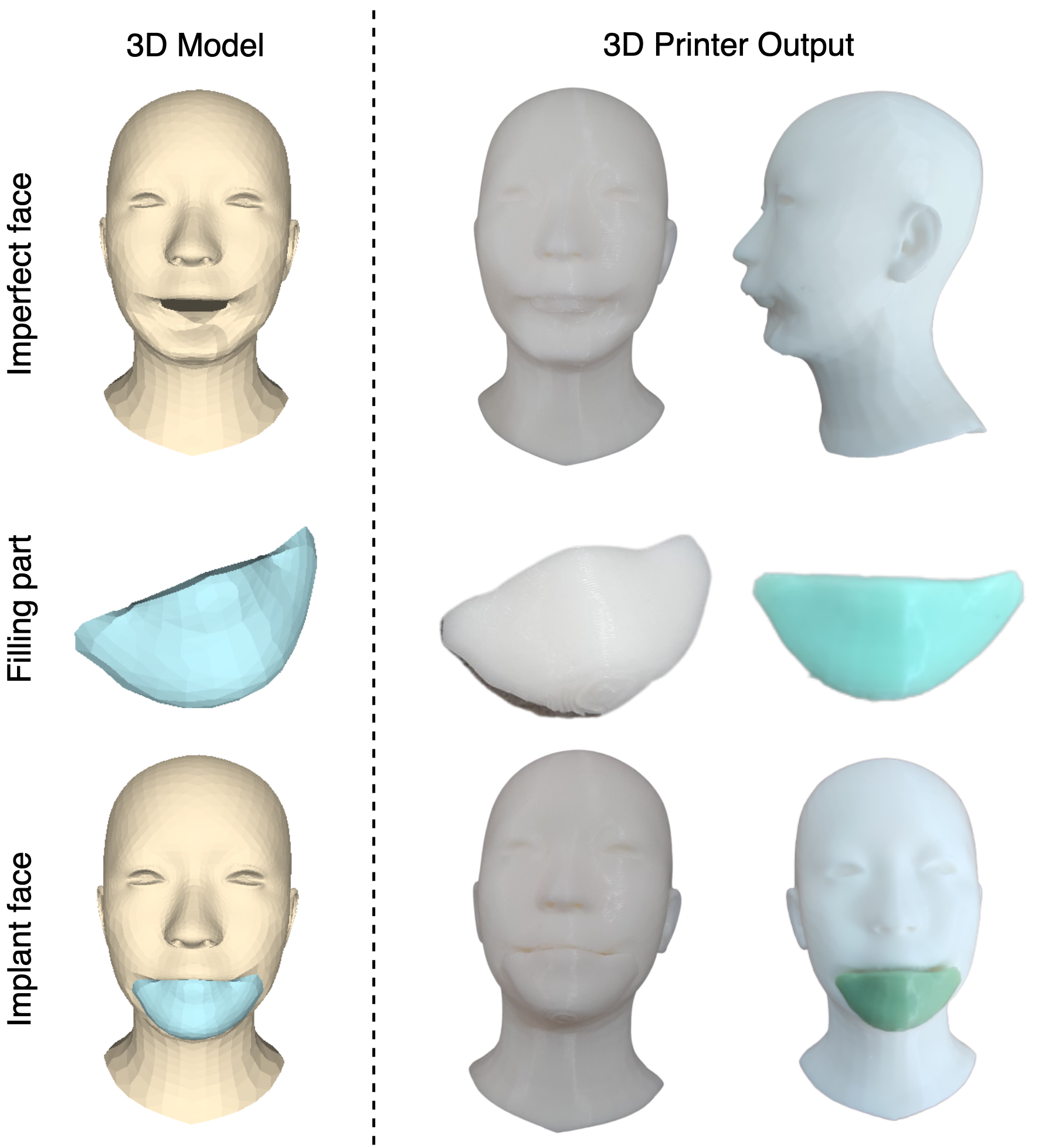}
	\caption{3D printer output from our model result.}
	\label{fig:3d_printing_output}
\end{figure}

The present research reveals the results of training the model to restore the facial structure when scarred in this part. The models were trained on the 2D-manifold training set and watertight mesh of the injured skull, as described in Section \ref{sec:model}. The scale used is that each human head mesh generates 10 human head meshes with wounds at random in a virtual space environment. Moreover, in the present approach, the visual model is likely to mirror the reality of individuals with facial injuries. Thus, the final dataset consists of 40458 human head mesh, including 3678 ground truth and 36780 input, derived from the starting dataset of 3687 human head mesh. As compared to previous studies, the current work consider the most hypothetical dataset of human faces available in the literature. The dataset generated by this work does not violate medical research ethics. In addition, it is also accessible to most models of injury in practice and those that are more severely harmed than they actually are. We employ 80\% of the aforesaid data as a training model, 10\% for validation, and the remaining 10\% for testing using this enormous data set. The facial selectivity model as shown in Fig. \ref{fig:illustration_for_the_middle_layer_of_the_model} uniformly divides the points in the middle layer, allowing the model to better replicate the facial parts with the figure trained with  $l_1$ reconstruction loss and Adam optimizer. We test the model on the test data and generate the statistics shown in Table \ref{tab:2}, and the error used is Geometric Distance. The results show that, while the single vertex error can be as high as 4.5643, the mean error of the overall vertices of mesh is only $2.75\times10^{-1}$. This demonstrates the correctness of the model in the overall vertices of the mesh in the majority of circumstances. Furthermore, this study may graphically represent the results displayed in Table \ref{tab:2} and Fig. \ref{fig:the_loss_function_value_corresponding_to_each_epoch}, where the chroma statistics of meshes displayed in Fig. \ref{fig:the_results_are_statistically_extracted_from_the_model} are derived based on the geometric distance of the vertices corresponding to output mesh and ground truth mesh. The research then randomly retrieved the outputs of the model on the test set, as shown in Fig. \ref{fig:the_results_were_randomly_extracted_from_the_model_with_the_test_suite}, to gain a better understanding of the outcomes. Simultaneously, the test employs the extraction of the filler, as shown in Fig. \ref{fig:3d_printing_output}, to proceed to the 3D printing procedure for wound restoration.

\begin{table}
\renewcommand{\arraystretch}{1.5}
\centering
\caption{Statistics extracted from the test suite (3678 meshes).}
\label{tab:2}
\begin{tabular}{|c|c|c|}
	\hline
Method	& Error  (geometric distance) & Figure reference \\
	\hline
Min vertices distance	& $2.26\times10^{-3}$ & Fig. \ref{fig:the_results_are_statistically_extracted_from_the_model}a \\
	\hline
Max vertices distance	& $4.564300$ & Fig. \ref{fig:the_results_are_statistically_extracted_from_the_model}b \\
	\hline
Mean vertices distance	& $2.75\times10^{-1}$ & Fig. \ref{fig:the_results_are_statistically_extracted_from_the_model}c \\
\hline
Min of Mean vertices distance	& $1.35\times10^{-1}$ & Fig. \ref{fig:the_results_are_statistically_extracted_from_the_model}d \\
\hline
Max of Mean vertices distance & $6.64\times10^{-1}$ & Fig. \ref{fig:the_results_are_statistically_extracted_from_the_model}e \\
\hline
\end{tabular}
\end{table}


\section{Conclusions}
This study explored the benefits of using the fully convolutional mesh autoencoder model for 3D object regeneration of individuals with facial trauma defects. The obtained results showed the most prominent features as follows
\begin{itemize}[leftmargin=0.3in]
\item[-] The virtual scar construction method allows the visual model in a virtual environment to accurately replicate the scars of patients with facial injuries. The research used a final dataset of 40,458 human head meshes, which is considered the most comprehensive human face dataset currently available. This dataset was created from an original set of 3,687 human head meshes and did not violate research ethics in medicine as it did not involve direct treatment of patients, but still provides access to a range of injury models and severe cases.
\item[-] The research results also support the proposed training method, which allows for the regeneration of facial scars based on the rest of the actual face. The model is able to accomplish this without mechanically reproducing scars or attaching one part of the person to others. The present work suggests using the outlier technique to separate the wound, which has been used to create a wound model for the entire face. The results obtained may provide useful insights for doctors during the re-surgery of patients' wounds. However, the model has limitations in its ability to accept a wide range of inputs, and future studies aim to address this issue in order to improve its practicality and overall strength. 
\item[-] The actual experiment results, as shown in Fig. \ref{fig:3d_printing_output}, indicate that 3D printing technology has the potential to improve the treatment process for patients. Firstly, it supports surgeons and patients in gaining a better understanding of the final results when the treatment process is completed. Secondly, this technology replaces traditional molding methods, enabling quicker mold sample creation and assisting surgeons in adapting to each stage of patient recovery. Finally, the combination of this research with bioprinting technology has the potential to create a filling part with biological materials that are not irritating to human skin and are implanted in the face to construct a natural-looking facial structure for patients. This is a promising development, as bioprinting technology allows for the creation of complex tissue structures that can be used for various medical applications such as drug testing, developing new therapies, and creating replacement body parts. These findings underscore the importance of ongoing research and development in the field of 3D printing technology and its potential to improve patient outcomes.
\item[-] A novel wound-covering model has been developed through this research, enabling patients to conceal facial imperfections. We hope this paper also leads to the creation of products that simplify and ensure the safe treatment of wounds during the healing process. The benefits of the current paper serve as a valuable aid for individuals suffering from illnesses in underprivileged, underdeveloped, and war-affected countries. The data generated from the present method is enriched and stored on our app system \url{https://ht3dprint.com}. 
\end{itemize}
\section*{Acknowledgments}
We would like to thank Vietnam Institute for
Advanced Study in Mathematics (VIASM) for hospitality during our visit in 2023, when we
started to work on this paper. It is important to highlight that the introduction of this article underwent refinement to enhance its readability, using a Large Language Model (LLM) called ChatGPT from OpenAI, based on our initial paper. Authors are accountable for the originality, validity, and integrity of the content of our submissions. The use of this AI-assisted technology is permitted by the 
\href{https://newsroom.taylorandfrancisgroup.com/taylor-francis-clarifies-the-responsible-use-of-ai-tools-in-academic-content-creation}{publisher's policy}.

\bibliographystyle{unsrt}  
\bibliography{3D_FaIR}

\end{document}